%% file: acl_latex.tex
\pdfoutput=1

\documentclass[11pt]{article}

\usepackage[]{acl}

\usepackage{times}
\usepackage{latexsym}

\usepackage[T1]{fontenc}

\usepackage[utf8]{inputenc}

\usepackage{microtype}

\usepackage{inconsolata}

\usepackage{cuted}
\usepackage{enumitem}
\usepackage{algorithm}
\usepackage{algpseudocode}
\usepackage{amsmath}
\usepackage{multirow}
\usepackage{booktabs}
\usepackage{url}
\usepackage{graphicx}
\usepackage{subcaption}
\usepackage{adjustbox}
\definecolor{RoseQuartzBg}{HTML}{F7CAC9}
\definecolor{RoseQuartz}{HTML}{F5A798}
\definecolor{Serenity}{HTML}{92A8D1}
\definecolor{OrangeRed}{rgb}{1.0, 0.27, 0.0}
\definecolor{Red}{rgb}{1.0, 0.0, 0.0}
\definecolor{Turquoise}{HTML}{0F4C81}
\definecolor{new_green}{HTML}{82B366}
\definecolor{new_red}{HTML}{B85450}
\usepackage{xparse}
\NewDocumentCommand{\lifu}{ mO{} }{\textcolor{OrangeRed}{\textsuperscript{\textit{Lifu}}\textsf{\textbf{\small[#1]}}}}
\NewDocumentCommand{\sijia}{ mO{} }{\textcolor{blue}{\textsuperscript{\textit{Sijia}}\textsf{\textbf{\small[#1]}}}}
\NewDocumentCommand{\pritika}{ mO{} }{\textcolor{purple}{\textsuperscript{\textit{changes}}\textsf{\textbf{\small[#1]}}}}
\NewDocumentCommand{\lalla}{ mO{} }{\textcolor{teal}{\textsuperscript{\textit{Lalla}}\textsf{\textbf{\small[#1]}}}}
\NewDocumentCommand{\joy}{ mO{} }{\textcolor{pink}{\textsuperscript{\textit{Joy}}\textsf{\textbf{\small[#1]}}}}
\usepackage{amssymb}
\usepackage{array}
\newcolumntype{L}{>{\centering\arraybackslash}m{3cm}}
\newcolumntype{S}{>{\centering\arraybackslash}m{2cm}}
\newcolumntype{P}{>{\arraybackslash}m{10cm}}
\newcolumntype{Q}{>{\arraybackslash}m{5cm}}

\makeatletter
\def\@fnsymbol#1{\ensuremath{\ifcase#1\or \dagger\or \ddagger\or
   \mathsection\or \mathparagraph\or \|\or **\or \dagger\dagger
   \or \ddagger\ddagger \else\@ctrerr\fi}}
    \makeatother

\usepackage{xspace}
\newcommand{\modelname}{\textsc{\textbf{RE}$^2$}\xspace}
\newcommand{\datasetname}{\textsc{\textbf{DiverseForm}}\xspace}

\title{\textsc{RE}$^2$: Region-Aware Relation Extraction from Visually Rich Documents}

\author{Pritika Ramu$^{\diamond}${\thanks{Work done while interning at Virginia Tech}} \quad Sijia Wang$^{\spadesuit}$ \quad Lalla Mouatadid$^{\clubsuit}$ \quad Joy Rimchala$^{\clubsuit}$ \quad Lifu Huang$^{\spadesuit}$ \\
  $^{\diamond}$ Adobe Research \quad
  $^{\spadesuit}$ Virginia Tech \quad
  $^{\clubsuit}$ Intuit AI Research \\
  \texttt{pramu@adobe.com} \quad \texttt{\{sijiawang,lifuh\}@vt.edu} \\ \texttt{\{lalla\_mouatadid,joy\_rimchala\}@intuit.com} \\}

\begin{document}
\maketitle
\begin{abstract}
Current research in form understanding predominantly relies on large pre-trained language models, necessitating extensive data for pre-training. However, the importance of layout structure (i.e., the spatial relationship between the entity blocks in the visually rich document) to relation extraction has been overlooked. In this paper, we propose \textbf{RE}gion-Aware \textbf{R}elation \textbf{E}xtraction (\modelname) that leverages region-level spatial structure among the entity blocks to improve their relation prediction. We design an edge-aware graph attention network to learn the interaction between entities while considering their spatial relationship defined by their region-level representations. We also introduce a constraint objective to regularize the model towards consistency with the inherent constraints of the relation extraction task. To support the research on relation extraction from visually rich documents and demonstrate the generalizability of \modelname, we build a new benchmark dataset, \textsc{DiverseForm}, that covers a wide range of domains. Extensive experiments on \textsc{DiverseForm} and several public benchmark datasets demonstrate significant superiority and transferability of \modelname across various domains and languages, with up to 18.88\% absolute F-score gain over all high-performing baselines\footnote{Code and dataset available at \url{https://github.com/VT-NLP/Form-Document-IE}}.
\end{abstract}

\input{sections/1intro}
\input{sections/2related_work}

\input{sections/4model}

\input{sections/5experiments}

\input{sections/6conclusion}

\input{sections/7Limitations}
\input{sections/8Ethics}

\section*{Acknowledgments}
This research is partially supported by a research award from Intuit AI Research and award No. 2238940 from the Faculty Early Career Development Program (CAREER) of the National Science Foundation (NSF). The views and conclusions contained herein are those of the authors and should not be interpreted as necessarily representing the official policies, either expressed or implied, of the U.S. Government. The U.S. Government is authorized to reproduce and distribute reprints for governmental purposes notwithstanding any copyright annotation therein.

\bibliography{custom}

\appendix

\input{sections/appendix}
\end{document}

%% file: sections/1intro.tex
\section{Introduction}
\label{sec:intro}

\begin{figure}[t!]
    \centering
    \includegraphics[width=\linewidth]{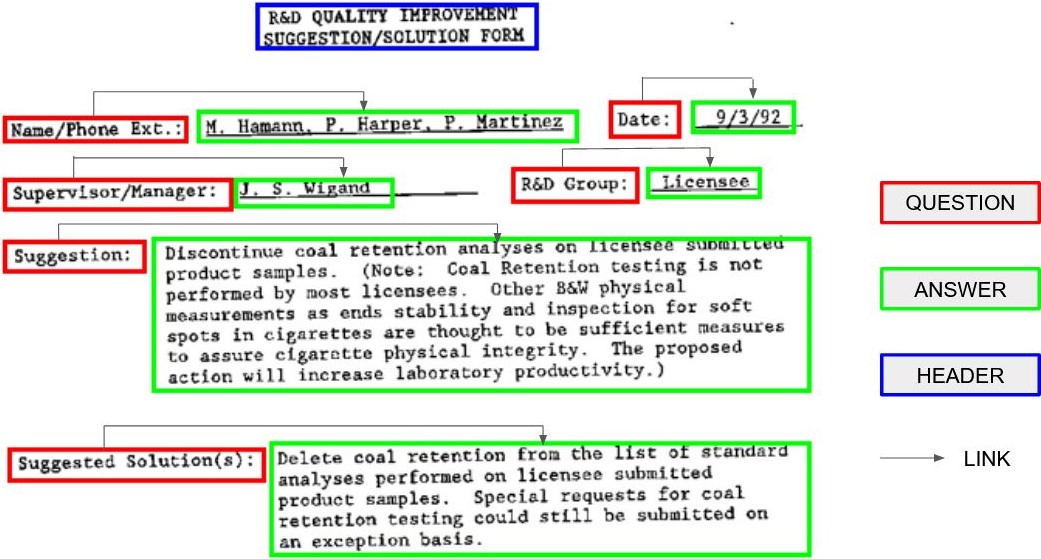}
    \vspace{-3mm}
    \caption{Example of entity and relation extraction from a visually rich document. The colored boxes represent three categories of semantic entities 
    and the arrows represent relations between them.
    \vspace{-5mm}
    }
    \label{fig:re_ex}
\end{figure}

Visually Rich Documents (VRDs) encompass various types such as \textit{invoices}, \textit{questionnaire forms}, \textit{financial forms}, \textit{legal documents}, and so on. These documents possess valuable layout information that aids in comprehending their content. Recent research \cite{liu2019graph,jaume2019funsd,yu2020pick} has focused on extracting key information, such as entities and relations, from VRDs by leveraging their layout structures and Optical Character Recognition (OCR) results\footnote{Optical Character Recognition will recognize a set of bounding boxes and their corresponding text from VRDs where each bounding box can represent a single word or a cohesive group of words, both semantically and spatially.}. 
Figure~\ref{fig:re_ex} shows an example where entity recognition aims to identify blocks of text in certain categories, such as \textit{Question}(\textit{Q}), \textit{Answer}(\textit{A}), and \textit{Header}(\textit{H}). Relation extraction further predicts the links among the entities, especially \textit{Q-A} links indicating that the \textit{A} block is the corresponding answer to the \textit{Q} block. 

Extracting key information, especially relations in VRDs is a challenging task. Though similar to traditional extraction tasks in text-only Natural Language Processing (NLP)~\cite{grishman1997information,chen2022new}, inferring relations in VRDs poses additional challenges. They require not only understanding the semantic meaning of entities but also taking into account the layout information, e.g., the spatial structures among the entity blocks in original VRDs. 
{Previous studies mainly focused on combining the text and layout with language model pre-training~\cite{ViLBERT,Su2020VL-BERT,chen2020uniter,powalski2021going_tilt,xu2022layoutlmv2, wang2022lilt, wang2022erniemmlayout, LayoutLMv3} or encoding the local layout information by constructing super-tokens~\cite{qian-etal-2019-graphie,liu2019graph,9412927,lee2022formnet, lee2023formnetv2}. However, the layout of the VRDs, especially the relative spatial relationship among the entity blocks, is still yet to be effectively explored for relation extraction.

To this end, we propose \textbf{RE}gion-Aware \textbf{R}elation \textbf{E}xtraction (\modelname) that leverages region-level spatial structures among the entities to reason about their relations\footnote{This work mainly focuses on extracting \textit{Q-A} relation given the gold \textit{Question} and \textit{Answer} entities.}. Specifically, given the question and answer entities from each VRD, we define three categories of region-level representations for each entity block, through which we further characterize the relative spatial relationship between each pair of question and answer entities. We then employ a layout-aware pre-trained language model (i.e., LayoutXLM~\cite{xu2022layoutlmv2}) to encode the entities and an Edge-aware Graph Attention Network (eGAT) to further learn the interaction between the question and answer entities in a bipartite graph 
while considering their spatial relationship. To ensure each answer is linked to at most one question, we design a constraint-based learning objective to guide the learning process, in combination with the relation classification objective.

To validate the effectiveness of \modelname{}, we conduct extensive experiments on various benchmark datasets for a wide range of languages and domains. We evaluate \modelname{} on two public datasets FUNSD~\cite{jaume2019funsd} and XFUND~\cite{xu-etal-2022-xfund}, under supervised, multitask transfer, and zero-shot cross-lingual transfer settings. We also create a new benchmark dataset \datasetname that covers diverse domains, such as Veterans Affairs, visa applications, tax documents, air transport and so on, and evaluate \modelname{} for cross-domain transfer. 
Experimental results show that \modelname{} outperforms the previous state-of-the-art approaches with a large margin on (almost) all languages and domains across all settings. Our ablation studies also verify the significant benefit of the region-level spatial structures of entity blocks for relation extraction. The contributions of this work are summarized as follows: 
\begin{itemize}
    \item We are the first to propose the region-level entity representations and utilize them to characterize the spatial structure among the entity blocks, which have been proven to be significantly beneficial to relation extraction from visually rich documents.

\vspace{-2mm}
    \item We develop a new framework \modelname{} that leverages the spatial structures among the question and answer entities with an effective eGAT network and regularizes model predictions with a novel constraint objective. \modelname{} demonstrates superior performance across (almost) all languages and domains under supervised, cross-lingual, and cross-domain transfer settings.
\vspace{-2mm} 
    \item We contribute \datasetname, a new benchmark dataset that covers a wide range of domains to support the research on information extraction from visually rich documents.
\vspace{-2mm}

\end{itemize}

%% file: sections/2related_work.tex
\section{Related Work}
\label{sec:related work}

\begin{figure*}[t]
    \centering
    \includegraphics[width=\linewidth]{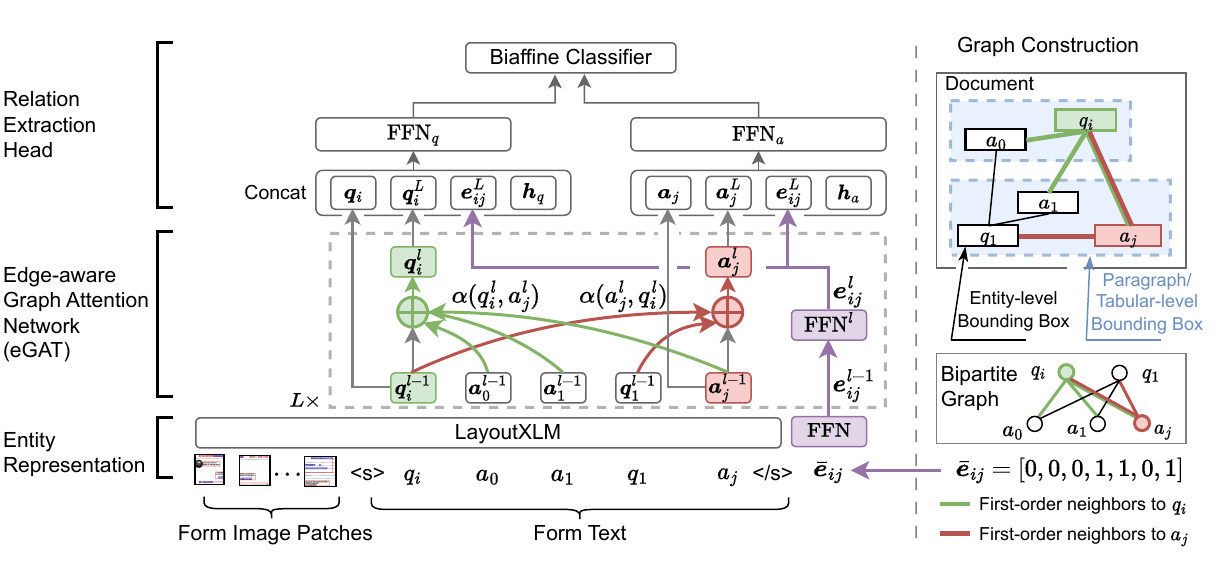}
    \vspace{-8mm}
    \caption{Overview of the \textbf{RE}gion-level \textbf{R}elation \textbf{E}xtraction (\modelname{}) framework. 
 {A bipartite graph of Question and Answer entities is constructed. In the eGAT layer, the representation of each entity is updated based on the attention scores of its first-order neighbors.} 
    }
    \vspace{-5mm}
    \label{fig:model}
\end{figure*}

Recent research on visually rich document information extraction shows that incorporating 2D positional embedding and layout coordinates into the pre-trained language models improves VRD understanding \cite{Xu_2020, xu2022layoutlmv2, LayoutLMv3, powalski2021going_tilt}.
\cite{wang2022erniemmlayout} models the spatial relationship of fine and coarse-grained visual elements based on Intersection over Union (IoU) and focuses only on named entity recognition task. Incorporating relative spatial positions of entities is essential for relation extraction task. \cite{luo2023geolayoutlm} incorporates the relative spatial relation between entities on a fine-grained level and serves as a task for model pre-training.
To deal with the variation of relation definitions, DocRel \cite{li2022relational} proposes a contrastive learning framework that 
utilizes the coherence of existing relations in diverse enhanced positive views to generate relation representations. 
\newcite{zhang2021entity} further explores entity relation extraction as dependency parsing, incorporating minimum vertical and horizontal distances between the entities as layout heuristics. Compared with all these studies, our approach is the first to propose and incorporate multi-granular spatial structures among the entities, which have been shown to significantly improve relation extraction from VRDs.

Graph Attention Networks (GAT) \cite{gat} have proven to be efficient for learning on graph-structured data~\cite{zhang-etal-2022-extracting}. 
This is exemplified by the work GraphDoc~\cite{zhang2022multimodal}, a multimodal graph attention-based model that simultaneously utilizes text, layout, and image information for visually rich document understanding. Though several studies~\cite{liu2019graph,lee2022formnet,lee2023formnetv2} have explored GNNs for entity extraction from VRDs, we are the first to design edge-aware GAT to improve relation extraction from VRDs, which presents additional challenges, encompassing spatial analysis to determine entity layout on the page and semantics between entities for identifying relations. 
GNNs have also been applied to relation extraction from textual documents~\cite{zhu-etal-2019-graph, guo-etal-2019-attention, zhang-etal-2018-graph}. However, these methods cannot be directly adapted to relation extraction from VRDs due to the fundamental differences in document formats, structures, and the key challenges encountered in relation extraction: text-only documents primarily rely on linguistic cues and phrases for relation extraction, whereas VRDs necessitate consideration of both semantics and spatial context. 
Given that, we innovatively incorporate a multi-granular layout heuristic into an edge-aware graph attention network, placing greater emphasis on capturing more fine-grained layout structures.

%% file: sections/4model.tex
\section{Approach}

Given a visually rich document $D$, 
a set of question entities $Q=\{q_1, q_2, ...,q_m\}$ and answers $A=\{a_1, a_2, ...,a_n\}$, we aim to identify all the connected pairs $(q, a)$ where $q\in Q$ and $a\in A$, indicating that $a$ is the corresponding answer of $q$. {Each $q_i$ or $a_j$ can be denoted as $\{[w_0, w_1,  \cdots, w_t], (x_0,y_0,x_1,y_1)\}$, where $[w_0, w_1, \cdots, w_t]$ is the sequence of words denoting the entity span and $(x_0,y_0,x_1,y_1)$ is the coordinates for the entity  bounding box. } 
Figure~\ref{fig:model} illustrates our \modelname framework that aims to leverage region-level spatial structures among the question and answer blocks to detect their association.

\subsection{Entity Representation}
\label{subsec:entity_encode}

We first learn the encoding of question and answer entities based on LayoutXLM~\citep{xu-etal-2022-xfund}, a layout-aware transformer-based model that has been extended to support multilingualism {by pretraining on multilingual VRD datasets}. 
 
{Given  a set of question entities $Q=\{q_1, q_2, ...,q_m\}$ and answers $A=\{a_1, a_2, ...,a_n\}$ from document $D$,}
we obtain the entity embeddings $\boldsymbol{Q}=\{\boldsymbol{q_1},\boldsymbol{q_2},...,\boldsymbol{q_m}\}$, $\boldsymbol{A}=\{\boldsymbol{a_1},\boldsymbol{a_2},...,\boldsymbol{a_n}\}$, $\boldsymbol{q_i,a_i} \in \mathbb{R}^\mathrm{1\times F}$, where $\mathrm{F}$ is the entity feature dimension\footnote{We use bold symbols to denote vectors.}. For entities with multiple tokens, we use the embedding of their first token as their representations\footnote{Preliminary experiments showed use of first subtoken performed better than average embedding of all subtokens.}. 

\begin{figure*}[t]
  \centering
  \begin{subfigure}{0.32\linewidth}
    \includegraphics[width=\linewidth]{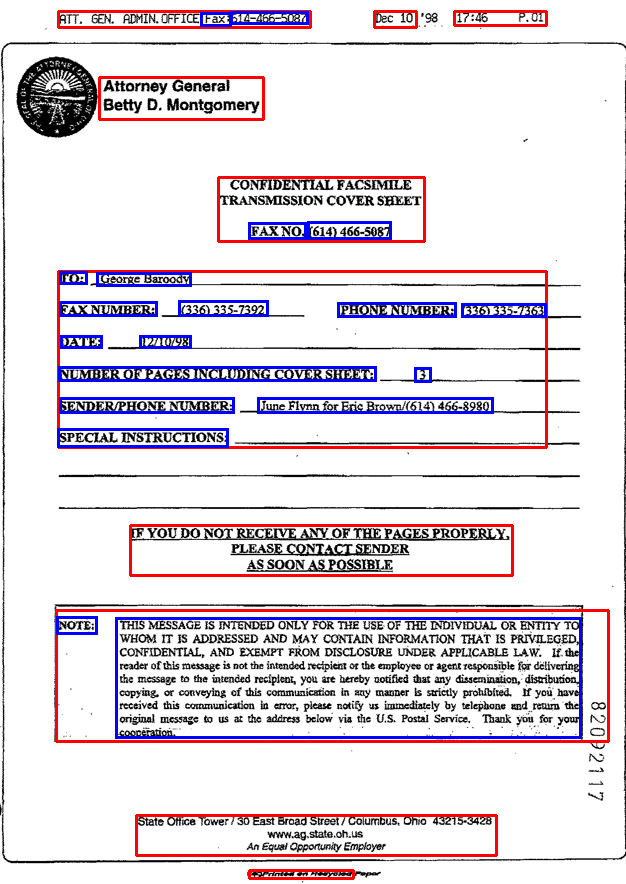}
    \caption{}
    \label{fig:regions_subfig1}
  \end{subfigure}
  \begin{subfigure}{0.32\linewidth}
      \includegraphics[width=\linewidth]{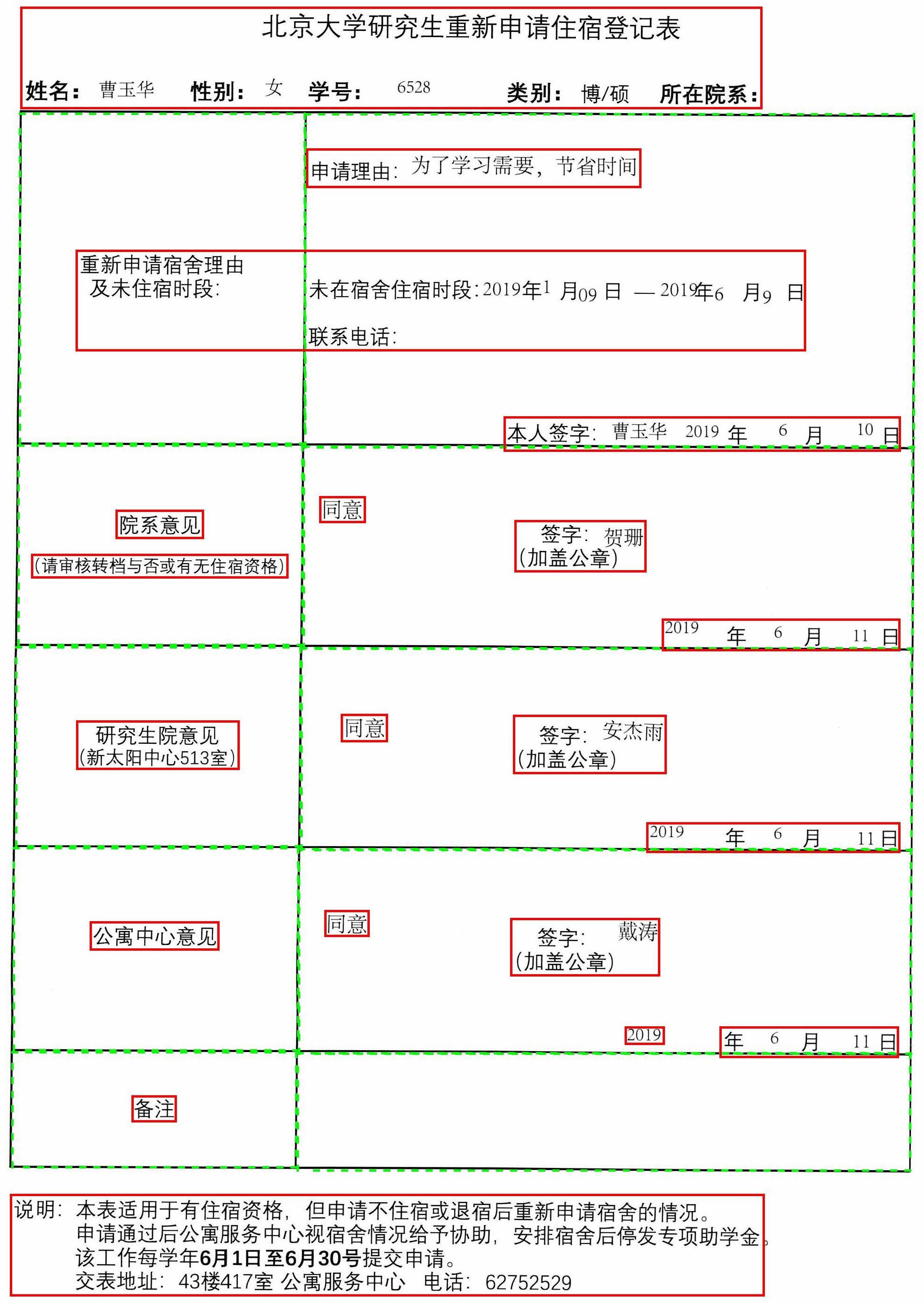}
      \caption{}
      \label{fig:regions_subfig2}
    \end{subfigure}
  \begin{subfigure}{0.34\linewidth}
      \includegraphics[width=\linewidth]{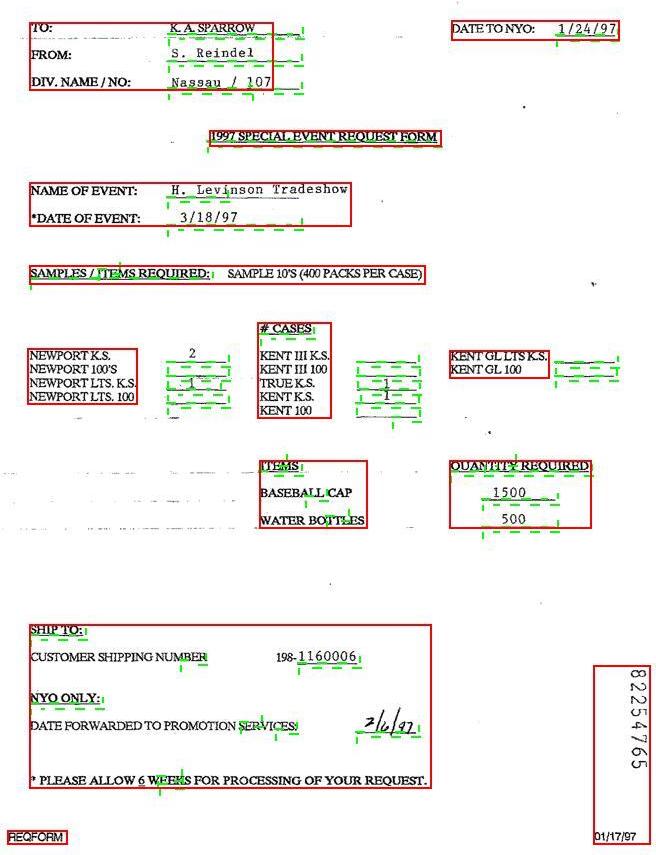}
      \caption{}
      \label{fig:regions_subfig3}
    \end{subfigure}
    \vspace{-2mm}
  \caption{Entity level bounding box (for question and answer entities) are shown in \textcolor{blue}{blue}, paragraph-level bounding box in \textcolor{red}{red} and tabular-based bounding box in \textcolor{green}{green}. 
  }
  \vspace{-4mm}
  \label{fig:regions}
\end{figure*}

\subsection{Region-Aware Graph Construction}

Based on the spatial structures of the input VRD, we define three distinct categories of regions (i.e., bounding box) for each entity: (1) an \textbf{entity-level bounding box} that refers to the bounding box encompassing the entire entity span and is obtained by merging the bounding boxes of all the words in a span obtained by OCR \cite{liu2019graph,yu2020pick}; 
(2) a \textbf{paragraph-level bounding box} that is defined as a visually distinct section for the paragraph where the entity occurs within a document and corresponds to the clustering of words that are located within a dense region. The paragraph-level bounding boxes are extracted by an existing tool, EasyOCR\footnote{\url{https://www.jaided.ai/easyocr/}}, which takes the maximum horizontal and vertical distances between adjacent word-level bounding boxes as hyperparameters to merge them into paragraph-level bounding boxes. {Other OCR systems include Tesseract \cite{Tesseract}, Microsoft OCR and other open source OCR systems provided by OpenCV\footnote{https://opencv.org}. Paragraph level bounding boxes can be obtained by clustering word level bounding boxes obtained from any of the OCR systems.};
and (3) a \textbf{tabular-based bounding box} if the entity occurs in a tabular structure demarcated by lines. We define a tabular-based bounding box as the coordinates of a table cell. Note that each entity can only appear in either a paragraph or a table, so other than its entity-level bounding box, we always assign either a paragraph-level or tabular-based bounding box for each entity, instead of both. 
{Our preliminary results show that a tabular-based bounding box is vital because tabular structures are usually not well-captured by existing OCR tools.}
Illustrations of the three types of regions are shown in Figure \ref{fig:regions}.
The pseudocode for extracting paragraph/tabular regions is present in Appendix \ref{sec:pseudocode}.

To characterize the links between the question and answer entities, we further propose to construct a complete bipartite graph, ${G}=(Q, A, E)$, for each visually rich document, where the question entities $Q=\{q_1, q_2, ...,q_m\}$ and answers $A=\{a_1, a_2, ...,a_n\}$ are the nodes, and for each pair of $q_i$ and $a_j$, there is an edge $e_{ij}\in E$ connecting them. Each entity is represented by the encoding learned from LayoutXLM as detailed in Section~\ref{subsec:entity_encode}, and each edge is represented by a one-hot encoding vector based on the spatial relationship between the three categories of bounding boxes of the question and answer: 
\begin{equation*}
\bar{\boldsymbol{e}}_{ij} = [\textrm I, \textrm E_{lr}^{1}, \textrm E_{tb}^{1}, \textrm E_{lr}^{0}, \textrm E_{tb}^{0}, \textrm R_{lr}, \textrm R_{tb}],
\end{equation*}
where each term is an indicator variable: $\textrm I$ indicates whether the two entities are within the same paragraph/tabular region. If so, $\textrm I=1$, otherwise, $\textrm I=0$. When the two entities are from the same paragraph/tabular region, $\textrm E_{lr}^{1}$ and $\textrm E_{tb}^{1}$
further indicate the left-right (lr) and top-bottom (tb) spatial relationship of their entity-level bounding boxes. For example, $\textrm E_{lr}^{1}=1$ indicates that the entity-level bounding boxes of the two entities have a left-right spatial relation, otherwise, $\textrm E_{lr}^{1}=0$. When the two entities are not from the same paragraph/tabular region, $\textrm E_{lr}^{0}$ and $\textrm E_{tb}^{0}$ indicate the left-right and top-bottom spatial relationship of their entity-level bounding boxes, while $\textrm R_{lr}$ and $\textrm R_{tb}$ indicate the left-right and top-bottom spatial relationship of their paragraph/tabular level bounding boxes. Note that when the two entities are from the same paragraph/tabular region, the indicators of $\textrm E_{lr}^{0}$, $\textrm E_{tb}^{0}$, $\textrm R_{lr}$, $\textrm R_{tb}$ will be all zero. A top-bottom relationship is defined based on the relative 
positions of the x coordinates ($[x_0, y_0, x_1, y_1]$ for $q_i$ and [$x_2, y_2, x_3, y_3$] for $a_j$). Specifically, a top-bottom relationship exists when either $x_0 \leq x_2 \leq x_1$, or $x_0 \leq x_3 \leq x_1$, or $x_2 \leq x_0 \leq x_3$, or $x_2 \leq x_1 \leq x_3$. Similarly, we define a left-right relationship based on the relative positions of the y coordinates, employing a similar logic. The intuition to determine the spatial relationship is to detect whether there is a vertical/horizontal overlap between region $q_i$ and $a_j$.

To obtain a dense representation of each edge, we pass each one-hot encoding vector $\bar{\boldsymbol{e}}_{ij}$ to a feed-forward network, and the resulting vector $\boldsymbol{e}_{ij} = \mathrm{FFN}(\bar{\boldsymbol{e}}_{ij})$ is assigned as the edge weight between $q_i$ and $a_j$, where $\boldsymbol{e}_{ij} \in \mathbb{R}^{1\times\mathrm{F}/2}$.

\subsection{Edge-aware Graph Attention Network}

We further propose an edge-aware graph attention network (eGAT)
, extended from the graph attention network (GAT)~\cite{gat} by incorporating the edge weights inferred by spatial information to learn the interaction between the question and answer nodes. In our experiments, eGAT consists of 2 encoding layers,  while each layer updates the node embeddings based on the first-order neighbors with masked self-attention. 

Specifically, given the node embeddings at layer $l$, $\boldsymbol{Q}^{l}=\{\boldsymbol{q_1^l},\boldsymbol{q_2^l},...,\boldsymbol{q_m^l}\}$, $\boldsymbol{A}=\{\boldsymbol{a_1^l},\boldsymbol{a_2^l},...,\boldsymbol{a_n^l}\}$, we first compute the attention weight between $q_i$ and $a_j$ as follows
\begin{equation*}
    \mathrm{att}(\boldsymbol{W}^l \boldsymbol{q}_i^l, \boldsymbol{W}^l \boldsymbol{a}_j^l)=\boldsymbol W_{att}^\top(\boldsymbol{W}^l \boldsymbol{q}_i^l|| \boldsymbol{W}^l \boldsymbol{a}_j^l)
\end{equation*}
\begin{equation*}
\boldsymbol{c}(q_i^l,a_j^l) = \text{LeakyReLu}\Big(\mathrm{att}\Big(\boldsymbol{W}^l \boldsymbol{q}_i^l, \boldsymbol{W}^l \boldsymbol{a}_j^l\Big)\Big)
\end{equation*}
\begin{equation*}
    \boldsymbol{\alpha}(q_i^l,a_j^l)=\underset{j}{\text{softmax}}\Big(\sum(\boldsymbol{e}_{ij}^{l}\cdot \boldsymbol{c}(\boldsymbol q_i^l,\boldsymbol a_j^l))\Big)
\end{equation*}
where $\cdot$ denotes scalar multiplication. $\boldsymbol{W}^l \in \mathbb{R}^{F^{\prime} \times F}$ is a parameter matrix for shared linear transformation for $\boldsymbol{q}_i^l$ and $\boldsymbol{a}_j^l$. 
$\boldsymbol{W}_{att}\in\mathbb R^{2F'}$ is a weight vector for the attention mechanism.
$||$ denotes the catenation operation. 
$\boldsymbol{e}_{ij}^{l} = \mathrm{FFN}^l(\boldsymbol{e}_{ij}^{l-1})$ where $\boldsymbol{e}_{ij}^{0}$ is the initial dense representation $\boldsymbol{e}_{ij}$ of each edge.

The resulting edge-aware normalized attention scores are then used to update the hidden representations of the question and answer nodes, respectively, with residual connection:
\begin{equation*}
 \boldsymbol{\bar{q}}_i^{l+1}=\boldsymbol{q}_i + \sum_{j \in \mathcal{N}_i} \boldsymbol{\alpha}(q_i^l,a_j^l) \boldsymbol{W} \boldsymbol{a}_j^l
\end{equation*}
\begin{equation*}
 \boldsymbol{\bar{a}}_j^{l+1}=\boldsymbol{a}_j + \sum_{i \in \mathcal{M}_j} \boldsymbol{\alpha}(a_j^l,q_i^l) \boldsymbol{W} \boldsymbol{q}_i^l 
\end{equation*}
where $\mathcal{N}_i$ and $\mathcal{M}_j$ denotes the first order neighbors of $q_i$ and $a_j$ respectively.

For each layer of eGAT, we apply multi-head attention~\cite{vaswani2017attention}, where each attention head performs operations independently, and the mean of all attention heads is taken for aggregation. The updated representation of question node $q_i$ and answer $a_j$ is computed as follows:
\begin{equation*}
\boldsymbol{q}_i^{l+1}=\sigma\bigg(\frac{1}{K} \sum_{k=1}^K \Big(\boldsymbol{q}_i +\sum_{j \in \mathcal{N}_i} \boldsymbol{\alpha}(q_i^l,a_j^l)^k \boldsymbol{W}^k \boldsymbol{a}_j^l\Big)\bigg)
\end{equation*}
\begin{equation*}
\boldsymbol{a}_j^{l+1}=\sigma\bigg(\frac{1}{K} \sum_{k=1}^K \Big(\boldsymbol{a}_j +\sum_{i \in \mathcal{M}_j} \boldsymbol{\alpha}(a_j^l,q_i^l)^k \boldsymbol{W}^k \boldsymbol{q}_i^l\Big)\bigg) 
\end{equation*}
where $K$ is the number of independent attention heads and $\boldsymbol{W}^k$ denotes the weight matrix for the $k^{th}$ attention head. $\sigma(\cdot)$ denotes a non-linear function (ELU is used for experiments). $\boldsymbol{q}_i^{l+1}$ and $\boldsymbol{a}_j^{l+1}$ are then used as input node embeddings for layer $l+1$.

\subsection{Relation Extraction}

\paragraph{Binary Relation Prediction} We predict a binary label for each question $q_i$ and answer $a_j$ pair, indicating their correspondence. The representations of $q_i$ and $a_j$ include LayoutXLM embedding, final node embedding from eGAT, edge representation of the pair, and an entity type representation (question or answer) learned by an embedding layer. The entity type embedding is crucial for determining the relation direction. The resulting $q_i$ and $a_j$ representations undergo two feed-forward networks and a biaffine classifier \cite{dozat2017deep} to obtain a score $s_{i,j}$ for determining the association between the pair.
\begin{equation*}
    \boldsymbol{q}_i^{'} = \mathrm{FFN_q}(\boldsymbol{q}_i\mathbin\Vert\boldsymbol{q}_i^{L}\mathbin\Vert \boldsymbol{e}^L_{ij}\mathbin\Vert \boldsymbol{h}_{q})
\end{equation*}
\begin{equation*}
    \boldsymbol{a}_j^{'} = \mathrm{FFN_a}(\boldsymbol{a}_j\mathbin\Vert\boldsymbol{a}_j^{L}\mathbin\Vert \boldsymbol{e}^L_{ij} \mathbin\Vert \boldsymbol{h}_{a})
\end{equation*}
\begin{equation*}
    {s}_{ij}=\boldsymbol{q}_i^{'}\boldsymbol{U} \boldsymbol a_j^{'}+\boldsymbol{V}\big(\boldsymbol{q}_i^{'} \circ \boldsymbol{a}_j^{'}\big)+\boldsymbol{b}
\end{equation*}
where $\boldsymbol{h}_{q}$ and $\boldsymbol{h}_{a}$ are the type embeddings of question and answer entities. Note that $\boldsymbol{h}_{q}$ and $\boldsymbol{h}_{a}$ remain the same across all questions and answers, respectively. $\boldsymbol{U}, \boldsymbol{V}$ and $\boldsymbol{b}$ are trainable parameters.
{During training, the loss is computed following the cross-entropy loss} 

\begin{equation*}
    \mathcal{L}_b = -\sum y\cdot\text{log}(p_{ij}).
\end{equation*}
where $y\in\{0,1\}$ is the target binary label and $p_{ij} = \text{softmax}(s_{ij})$, indicating the probability of  a relation between $q_i$ and $a_j$.

\begin{table*}[ht]
  \centering
  \small
  \begin{tabular}{l|c|c|c|c|c|c|c|c|c}
    \toprule
    Model & EN & ZH & JA & ES & FR & IT & DE & PT & Avg.\\
    \midrule
    XLM-RoBERTa$_\text{BASE} $~\cite{conneau-etal-2020-unsupervised} & 26.59 & 51.05 & 58.00 & 52.95 & 49.65 & 53.05 & 50.41 & 39.82 & 47.69\\
    InfoXLM$_\text{BASE}$~\cite{chi-etal-2021-infoxlm} & 29.20 & 52.14 & 60.00 & 55.16 & 49.13 & 52.81 & 52.62 & 41.70 & 49.10\\
    LayoutXLM$_\text{BASE}$~\cite{xu-etal-2022-xfund} & 54.83 & 70.73 & 69.63 & 68.96 & 63.53 & 64.15 & 65.51 & 57.18 & 64.32\\
    LiLT[InfoXLM]$_\text{BASE}$~\cite{wang2022lilt} & 62.76 & 72.97 & 70.37 & 71.95 & 69.65 & 70.43 & 65.58 & 58.74 & 67.81 \\
\midrule
    \modelname (Our Approach) & \textbf{71.76} & \textbf{79.60} & \textbf{75.36} & \textbf{75.59} & \textbf{76.38} & \textbf{77.45} & \textbf{75.86} & \textbf{59.76} & \textbf{73.98}\\
    \bottomrule
  \end{tabular}
  \vspace{-2mm}
  \caption{Language-specific fine-tuning results (F1\%) on FUNSD(EN) and XFUND.}
  \label{tab:results1}
\end{table*}

\begin{table*}[ht]
  \centering
  \small
  \begin{tabular}{l|c|c|c|c|c|c|c|c|c}
    \toprule
    Model & EN & ZH & JA & ES & FR & IT & DE & PT & Avg.\\
    \midrule
     XLM-RoBERTa$_\text{BASE}$ & 26.59 & 16.01 & 26.11 & 24.40 & 22.40 & 23.74 & 22.88 & 19.96 & 22.76 \\
     InfoXLM$_\text{BASE}$ & 29.20 & 24.05 & 28.51 & 24.81 & 24.54 & 21.93 & 20.27 & 20.49 & 24.23 \\
    LayoutXLM$_\text{BASE}$ & 54.83 & 44.94 & 44.08 & 47.08 & 44.16 & 40.90 & 38.20 & 36.85 & 43.88\\
    LiLT[InfoXLM]$_\text{BASE}$ & 62.76 & 47.64 & 50.81 & 49.68 & 52.09 & 46.97 & 41.69 & 42.72 & 49.30 \\\midrule
    \modelname (Our Approach) & \textbf{71.76} & \textbf{66.32} & \textbf{64.42} & \textbf{58.82} & \textbf{69.02} & \textbf{61.83} & \textbf{60.57} & \textbf{43.87} & \textbf{62.08}\\
    \bottomrule
  \end{tabular}
  \vspace{-2mm}
  \caption{Zero-shot cross-lingual results (F1\%) (trained on EN (FUNSD) and tested on other languages)}.
  \vspace{-6mm}
  \label{tab:results2}
\end{table*}

\paragraph{Constraint Loss}
Our preliminary study shows that without any constraint, the model tends to predict multiple questions to be associated with one answer, which is against the definition of relation extraction for VRDs, where each answer is linked to at most one question. 
To address this issue, we incorporate the constraint into the learning process in the form of a constraint loss.
Previous work~\cite{li-etal-2019-logic, wang-etal-2020-joint} demonstrated that declarative logical constraints can be converted into differentiable functions, and help regularize the model towards consistency with the logical constraints. We design a declarative logical constraint that holds true for relation extraction task from VRDs as follows,
 $\forall a_j \in {A}, \;\forall q_i \in {Q}$, 
\begin{align*}
    \text{rel}(q_i,a_j)\rightarrow  \bigwedge_{q_k\in\textbf{Q}\setminus\{q_i\}} \neg\;\text{rel}\;(q_k,a_j).
\end{align*}
This means, for any $a_j\in A$, if there exists one relation link between $a_j$ and any particular $q_i$ among all questions, there cannot be another relation link for this answer $a_j$. 
We further define the following constraint loss derived from the logical constraints:
\begin{equation*}
    \mathcal{L}_c = y \cdot \bigg|\log(p_{ij}) - \frac{1}{|Q| - 1}\sum_{\substack{k=0 \\ k \neq i}}^{|Q|} \log\left(1 - p_{kj}\right)\bigg|
\end{equation*}
where $Q$ denotes the whole set of questions in the document. 

\paragraph{Overall Learning Objective}
The overall learning objective is a weighted combination of the binary cross entropy loss and the constraint loss:
\begin{equation*}
       \mathcal{L} =  \beta \mathcal{L}_b + \delta \mathcal{L}_c
\end{equation*}
where $\beta$ and $\delta$ are hyperparameters.

%% file: sections/5experiments.tex
\section{Experiment Settings}

\subsection{Datasets}
The primary challenge in relation extraction from visually rich documents is the diverse layouts in form-like documents across domains and languages. However, \modelname addresses this by introducing domain and language-independent region-level spatial structures. To validate its effectiveness, we conduct experiments on diverse benchmark datasets spanning multiple languages and domains.

\paragraph{FUNSD}
The FUNSD dataset \cite{jaume2019funsd} is derived from the RVL-CDIP dataset \cite{harley2015evaluation}, featuring scanned document images with OCR ground truth. It includes bounding boxes and annotations for four entity types: \textit{Question}, \textit{Answer}, \textit{Header}, and \textit{Other}. The dataset emphasizes relational links, particularly focusing on Question-Answer links. We follow the data split and experimental settings of prior studies \cite{xu-etal-2022-xfund, wang2022lilt}, utilizing 149 documents for training and 50 for evaluation, and report the best performance on the evaluation set.

\begin{table*}[ht]
  \centering
  \small
  \begin{tabular}{l|c|c|c|c|c|c|c|c|c}
    \toprule
    Model & EN & ZH & JA & ES & FR & IT & DE & PT & Avg.\\
    \midrule
    XLM-RoBERTa$_\text{BASE}$ & 36.38 & 67.97 & 68.29 & 68.28 & 67.27 & 69.37 & 68.87 & 60.82 & 63.41 \\
    InfoXLM$_\text{BASE}$ & 36.99 & 64.93 & 64.73 & 68.28 & 68.31 & 66.90 & 63.84 & 57.63 & 61.45 \\
    LayoutXLM$_\text{BASE}$ & 66.71 & 82.41 & 81.42 & 81.04 & 82.21 & 83.10 & 78.54 & 70.44 & 78.23 \\
    LiLT[InfoXLM]$_\text{BASE}$ & 74.07 & 84.71 & \textbf{83.45} & \textbf{83.35} & 84.66 & \textbf{84.58} & 78.78 & \textbf{76.43} & 81.25 \\ \midrule
    \modelname (Our Approach) & \textbf{74.11} & \textbf{88.25} & 82.27 & 83.23 & \textbf{86.83} & 84.02 & \textbf{81.89} & 71.04 & \textbf{81.46}\\
    \bottomrule
  \end{tabular}
  \vspace{-2mm}
  \caption{ Multitask fine-tuning performance (F1\%) on FUNSD(EN) and XFUND.}
  \label{tab:results3}
\end{table*}

\begin{table*}[ht]
\centering
\small
\begin{tabular}{l|c|c|c}
\toprule

     Model & \datasetname & FUNSD $\rightarrow$ \datasetname & \datasetname $\rightarrow$ FUNSD \\    \midrule
    LayoutXLM$_\text{BASE}$ & 69.72 & 37.33 & 32.58\\
    LiLT[InfoXLM]$_\text{BASE}$ & 64.15 & 41.56 & 30.26 \\
    \modelname & \textbf{70.87} & \textbf{41.78} & \textbf{50.32}\\
\bottomrule
\end{tabular}
\vspace{-2mm}
\caption{Supervised results on \datasetname and cross-domain transfer results between \datasetname and FUNSD. (F1\%)}
\label{tab:results custom data}
\vspace{-4mm}
\end{table*}

\begin{table*}[ht]
\centering
\small
\begin{tabular}{l|c|c|c|c|c|c|c|c|c}
\toprule
     Model & EN & ZH & JA & ES & FR & IT & DE & PT & Avg.\\    \midrule
\modelname & \textbf{71.76} & \textbf{79.60} & \textbf{75.36} & \textbf{75.59} & \textbf{76.38} & \textbf{77.45} & \textbf{75.86} & \textbf{59.76} & \textbf{73.98}
\\\midrule
 \quad - node embedding  & 70.19 & 78.93 & 75.00 & 74.60 & 76.00 & 76.82 & 73.20 & 57.29 & 72.75 \\ 
 \quad - edge embedding  & 57.42  & 69.37 & 67.93 & 72.01 & 73.73 & 69.67 & 63.48 & 55.61 & 66.15\\ \midrule 
 \quad - constraint loss & 68.52 & 77.77 & 74.49 & 74.78 & 75.20 & 75.66 & 73.61 & 57.48 & 72.19\\ \midrule
\quad - entity level regions & 44.69 & 76.89 & 66.71 & 73.11 & 62.44 & 70.63 & 62.10 & 44.30 & 62.61\\
\quad - paragraph/tabular regions & 71.57 & 79.5 & 74.17 & 72.05 & 74.98 & 76.79 & 74.55 & 57.49 & 72.64\\ 
\bottomrule
\end{tabular}
\vspace{-2mm}
\caption{Ablation study results (F1\%) on eGAT (node and edge embeddings), constraint loss, paragraph/tabular regions and entity level regions.}
\vspace{-2mm}
\label{tab:abl1}
\end{table*}

\paragraph{XFUND}
XFUND \cite{xu-etal-2022-xfund} is a diverse multilingual dataset with visually rich documents in seven languages: Portuguese, Chinese, Spanish, French, Japanese, Italian, and German. Featuring 1,393 fully annotated forms, each language has 149 forms for training and 50 for testing, providing ground truth OCR, entity, and relation annotations. Notably, XFUND shares document format similarities with the FUNSD dataset.

\paragraph{\datasetname}
To best demonstrate the performance of domain transfer of \modelname{}, we further create a new dataset, \datasetname, by curating government forms from~\newcite{aggarwal2020multi} and~\newcite{sarkar2020document}. These forms encompass a wide range of question types, including checkboxes, tables, multiple-choice questions (MCQs), and fill-in-the-blank fields. 
The domains of the forms cover various areas such as Veterans Affairs, visa applications, tax documents, air transport, legal forms, vehicle-related forms from the Department of Motor Vehicles (DMV), and miscellaneous forms from different government agencies. 
These forms are of single page and were originally empty and they are designed to collect confidential information such as health data and tax details. To populate the forms, we employed two annotators who used synthetic data generated by The One Generator\footnote{\url{https://theonegenerator.com/}} for fields such as names, addresses, and other necessary information.
This approach ensures the privacy and security of individuals' personal information while providing a realistic representation of the data typically found in these government forms. We then hire another annotator to label the \textit{Question} and \textit{Answer} entities as well as their relations for these documents using the annotation tool UBIAI\footnote{\url{https://ubiai.tools/}}, which also offers its customized OCR model for extracting text from uploaded images. 
However, due to the serialized top-left to bottom-right text extraction approach of the OCR, the spans of entities are sometimes fragmented in complex layout forms. During the annotation process, these fragmented spans are identified and merged to achieve the correct serialization of spans. After labeling the entities and relations for these documents, we further hire three annotators to validate the annotations. All the annotators are senior undergraduate students majoring in Computer Science and are paid a rate of $\$15/hour$. 
We name the final annotated dataset as \datasetname, which comprises a total of 150 training documents and 50 testing documents. 
Details of \datasetname annotation and statistics is in Appendix \ref{sec:annotation_detail}.

\section{Experiment Setting and Hyperparameters}
The NVIDIA A40 GPU was utilized for all fine-tuning tasks. Paragraph-level regions are created using EasyOCR through horizontal merging of text boxes when their distance is within 2, and vertical merging is performed when the distance is within 1, with the paragraph flag set to True. The model undergoes end-to-end training, incorporating fine-tuning of the LayoutXLM base model. The eGAT layers and relation extraction head are trained from scratch, employing 2 eGAT layers for all experiments. The training process consists of 5000 steps with a batch size of 4, a learning rate of 5e-5, and a warm-up ratio of 0.1. Cross-entropy loss is weighted at 1, and constraint loss is weighted at 0.02.

\subsection{Inference Details}
During the inference phase, the input comprises head entities, tail entities, bounding boxes (acquired from OCR), and the document image. This input undergoes a modeling process similar to the training phase, wherein additional processing is applied to derive entity-level, paragraph-level, and tabular-level bounding boxes. Subsequently, these bounding boxes are normalized to establish a relative spatial representation of entities, facilitating relation extraction tasks.

\subsection{Experiment Results}

\textbf{Language-specific fine-tuning} results are presented in Table \ref{tab:results1}, where each model is fine-tuned on language X and tested on language X. The experimental findings show that the proposed model outperforms all the baselines across all evaluated languages.
To evaluate the \textbf{cross-lingual zero-shot transfer} capability, the model is fine-tuned on the FUNSD dataset in English, followed by testing on multiple languages. The experimental results, as shown in Table \ref{tab:results2}, demonstrate the superiority of our model over the baseline approach in terms of zero-shot performance. This outcome provides compelling evidence that the incorporated region-level spatial structures and constraints for relation extraction exhibit effective transferability across different languages. We also conduct a significance test for both our approach and the best-performing baseline (i.e., LiLT[InfoXLM]$_\text{BASE}$~\cite{wang2022lilt}) under the settings of language-specific fine-tuning and cross-lingual zero-shot transfer. As shown in Table \ref{tab:p_val} in Appendix \ref{sec:appendix_p_val}, our approach significantly outperforms the baseline under both settings.
Table \ref{tab:results3} displays the results of \textbf{multitask fine-tuning}, where the model is trained on all language training sets and tested on each individual language. The superior performance showcases the model's successful learning of layout invariance across languages. By capturing shared layout characteristics, the model demonstrates improved generalization, enhancing performance across diverse linguistic contexts. This emphasizes the importance of incorporating layout information in cross-lingual settings and underscores the model's adaptability and knowledge transfer for effective document processing across various languages.

Note that \modelname shows less competitive performance on Portuguese (PT) due to more complex layout structures. Portuguese forms exhibit a combination of mixed tables and paragraph structures, making it challenging to determine the appropriate usage for paragraph-level regions or tabular regions. An example is shown in Appendix \ref{sec:appendix_pt}.

We also assess the generalization of \modelname and two high-performing baselines based on \datasetname and FUNSD, which cover two sets of distinct domains. 
We conduct experiments under the settings of both domain-specific fine-tuning and cross-domain transfer where the models are trained on one dataset and tested on the other. As shown in Table~\ref{tab:results custom data}, \modelname significantly outperforms the two strong baselines when fine-tuned on \datasetname and tested on \datasetname or FUNSD. The improvement of \modelname when it's trained on FUNSD and tested on \datasetname is marginal, probably due to the greater diversity and complexity in document layout of \datasetname compared to FUNSD.

\subsection{Ablation Study}
\paragraph{Effect of Node and Edge Embeddings from eGAT}
The node and edge embeddings from eGAT are concatenated with the entity representations before being passed to the biaffine classifier. A series of ablation studies are conducted to assess the individual contributions of the layout information. The results of these studies are presented in Table \ref{tab:abl1}. Figure \ref{fig:abl1} in Appendix \ref{sec:viz_ablation} provides visual evidence that solely relying on the updated node embeddings from eGAT fails to adequately capture the layout heuristics and results in the omission of numerous relations. Conversely, employing only the updated edge embeddings without considering the node embeddings leads to an over-prediction of relations with limited regard for the semantic relevance of the entities involved. Optimal performance is achieved through the joint utilization of both node and edge embeddings, indicating the importance of integrating both sources of information to effectively capture the region-level spatial structures and consider the semantic context of the relations.

\paragraph{Effect of Constraint Loss}
The constraint loss has been modeled to encourage each answer entity to be linked to at most one question. Table~\ref{tab:abl1} shows that incorporating the constraint loss significantly improves the F1 score of \modelname, especially precision. The detailed experimental results are evidenced in Appendix \ref{sec:ablation}.

\paragraph{Effect of Region Information}

We also investigate the impact of each category of regions on characterizing the spatial relationship among the entities and further affecting the performance of \modelname. As shown in Table~\ref{tab:abl1}, the inclusion of each category of region information significantly improves the performance of \modelname. 
The absence of entity-level regions resulted in a substantial decrease in performance, underscoring the vital role of pairwise entity layout information, i.e., whether the question and answer entities are arranged vertically (top-bottom) or horizontally (left-right).

Figure \ref{fig:abl2} in Appendix \ref{sec:viz_ablation} shows an example to compare the relation predictions with and without paragraph/tabular regions, indicating that
incorporating paragraph/tabular regions helps prevent the model from predicting relations across semantically different regions. The result of this ablation study proves the effectiveness of the multi-granular region information.

%% file: sections/6conclusion.tex
\section{Conclusion}
In this work, we propose a novel entity relation extraction model, \modelname, that incorporates layout heuristics and constraints that are generalizable across different languages. Experimental results on 8 different languages {and our proposed dataset \datasetname} show the effectiveness of our proposed method under four settings (language-specific, cross-lingual zero-shot, multi-lingual fine-tuning, and cross-domain transfer).

%% file: sections/7Limitations.tex
\section*{Limitations}
In this work, we found the incorporation of layout heuristics to be compelling and we are excited by how leveraging region information improves performance drastically. 
One of the limitations of our model is its reliance on a relatively limited set of heuristics and features. For instance, we have not yet incorporated visual information and template-based knowledge, which could potentially improve the accuracy and robustness of the relation extraction task. Additionally, the current model employs an exhaustive inference approach, considering all possible relations during prediction. While this ensures comprehensive coverage, it also results in longer inference times for each relation type. These limitations indicate avenues for further improvement, such as exploring additional heuristics and incorporating more efficient inference strategies, to enhance the performance and efficiency of our model.

%% file: sections/8Ethics.tex
\section*{Ethical Considerations}
The forms in the DiverseForm dataset are synthetically constructed and should not be mistaken for real forms. The values within these forms are populated through random generation, adhering to patterns that reflect typical data; however, these entries are not genuine. By employing synthetic data, we ensure that the model is trained on data closely resembling real-world scenarios without compromising the privacy and security of actual individuals. This approach is in line with ethical guidelines that prioritize data protection and privacy rights, making it a responsible choice for developing models that handle sensitive information. The proposed model is designed to enhance understanding of various document layouts, including checkboxes, tables, and fill-in-the-blank fields—areas often overlooked in previous studies. Its potential misuse is dependent on unauthorized access to genuine information.

%% file: sections/appendix.tex
\section*{Appendix}
\label{sec:appendix}
\section{Data Preprocessing}
To accurately determine the layout heuristics, it is important to get the bounding box of the entire entity span. If token-level bounding boxes are provided, the boxes can be merged to obtain a span-level box. All the paragraph/tabular regions are detected and their bounding boxes are obtained. We identify the region an entity belongs to by checking the entity's Intersection over Union (IoU) with the regions and assign the region with the maximum IoU.

\section{Annotation Details and Statistics of \datasetname}
The guidelines of annotating entities are as follows:
\begin{itemize}
    \item Question: A word, set of words, or sentence worded or expressed so as to elicit information from the person filling the form.
    \begin{itemize}
        \item Questions are annotated even if they haven’t been answered
        \item Questions and sub-questions are labeled as the same type of entity.
    \end{itemize}
    \item Header: A word, set of words, or sentences worded or expressed so as give context or encapsulate a set of questions.
    \begin{itemize}
        \item Annotate headers even if their questions haven’t been answered
        \item Headers do not have answers directly attached to them
    \end{itemize}
    \item Answer: A word, set of words, or sentence written in response to a question. 
    \begin{itemize}
        \item Responses in the form of checkbox options count as answers.
        \item In multiple choice type questions, all the options are annotated as answers (following FUNSD and XFUND)
    \end{itemize}
\end{itemize}
The guidelines of annotating relations are as follows:
\begin{itemize}
    \item Question-Answer: A link exists between a question entity and an answering entity when the answer is a response to a particular question.
    \begin{itemize}
        \item When multiple answers exist for a question, there are multiple Question-Answer links from the same question entity.
        \item Answers to a sub-question should only be linked to the sub-question and not the parent question.
    \end{itemize}
    \item Question-Question: A link exists between a question entity and another question entity if one question is a sub-question of another question or one question is conditioned on the answer of another question.
    \begin{itemize}
        \item For example, “If yes, …” type of question has a Question-Question link with the parent question.
        \item A question that is split into multiple fine-grained questions has a Question-Question link between them. For example, "Address" can have further questions such as "Apt. No", "Street Name", "City", "State", "Zip Code".
    \end{itemize}
    \item Header-Question: A link exists between a header entity and a question entity if the questions are present under the section or subsection that is characterized by the header.
    \begin{itemize}
        \item If multiple questions exist under a header, there are multiple Header-Question links from the same header entity.
        \item Often confused with Question-Question links and can be differentiated based on layout structure, font style, and other visual aspects of the questions from the form.
    \end{itemize}
\end{itemize}

The guidelines of annotation of tables are as follows:
We mainly deal with one dimensional tables. For the case that each cell in the table is related to both row and column questions, there will be a Question-Question link between the questions extracted from the row and column, indicating that one question is a sub-question of another question or one question is conditioned on the answer of another question. This is part of the annotation guidelines for FUNSD and our own dataset. Based on these annotation rules, the constraint of one answer having one question still holds.

\begin{figure*}[t]
    \centering
    \includegraphics[width=0.7\linewidth]{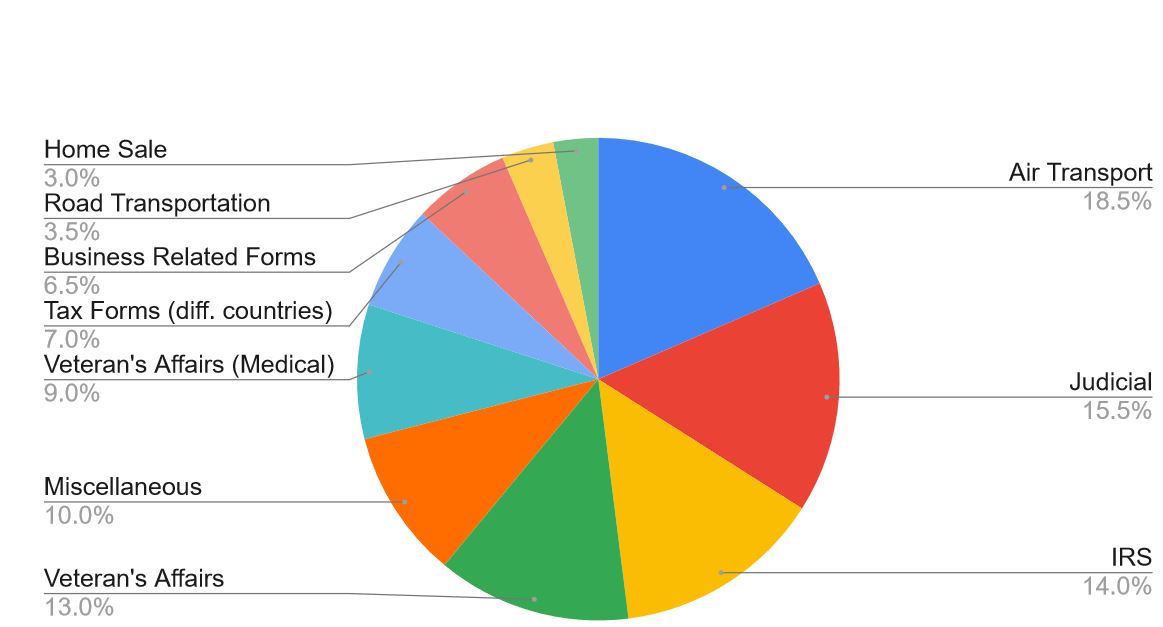}
    \vspace{4mm}
    \caption{Domain distribution of \datasetname.}
    \label{fig:chart}
\end{figure*}

Figure \ref{fig:chart} shows the distribution of domains in \datasetname. Miscellaneous consists of forms for voter registration, agriculture, scholarship, immigration, property tax, etc. Veteran's Affairs encompasses varying forms ranging from child support payments to retirement funds. There is rich layout variation within each domain shown in the chart. The number of entities and relations of each type in \datasetname are tabulated in Table \ref{tab:stats}.

\begin{table*}[ht]
  \centering
  \small
    \centering
    \begin{tabular}{c|c|c|c|c|c|c}
    \toprule
        & \multicolumn{3}{c|}{Entities} & \multicolumn{3}{c}{Relations} \\
        \midrule
        Split & Question & Answer & Header & Question-Answer & Question-Question & Header-Question\\
        \midrule
         Training & 3,087 & 3,585 & 230 & 1,172 & 594 & 546\\
         Test & 956 & 1,048 & 57 & 520 & 270 & 164\\
         \bottomrule
    \end{tabular}
    \caption{Statistics of entities and relations in \datasetname}
    \label{tab:stats}
\end{table*}

\label{sec:annotation_detail}

\section {Significance Test}
\label{sec:appendix_p_val}
Table~\ref{tab:p_val} shows the significance test results for both our approach and the best performing baseline (i.e., LiLT[InfoXLM]$_\text{BASE}$~\cite{wang2022lilt}) under the settings of language-specific fine-tuning and cross-lingual zero-shot transfer. The results for all experiments reported were averaged across 3 runs.
\begin{table*}[ht]
  \centering
  \begin{tabular}{l|c|c|c|c|c}
    \toprule
     & \multicolumn{2}{c|}{\modelname} & \multicolumn{2}{c|}{Baseline} &  \\
     \midrule
    Setting & Mean & SD & Mean & SD & P-value \\
    \midrule
    Language-Specific Fine-Tuning & 73.98 & 6.15 & 67.81 & 4.98 & 0.0447 \\
    Zero-Shot Cross-Lingual & 62.08 & 8.53 & 49.30 & 6.55 & 0.0047 \\
    \bottomrule
  \end{tabular}
  \caption{Significance Test Results}
  \label{tab:p_val}
\end{table*}

\section{Case Study}
\label{sec:appendix_pt}
\begin{figure*}[!htp]
  \centering
  \begin{subfigure}{0.45\linewidth}
  \centering
    \includegraphics[width=\linewidth]{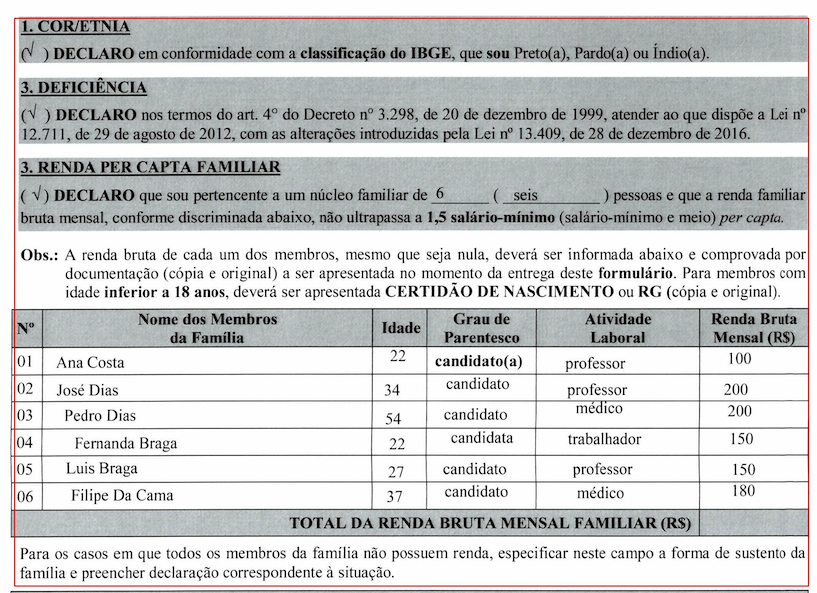}
    \caption{}
  \end{subfigure}
  \begin{subfigure}{0.45\linewidth}
  \centering
    \includegraphics[width=\linewidth]{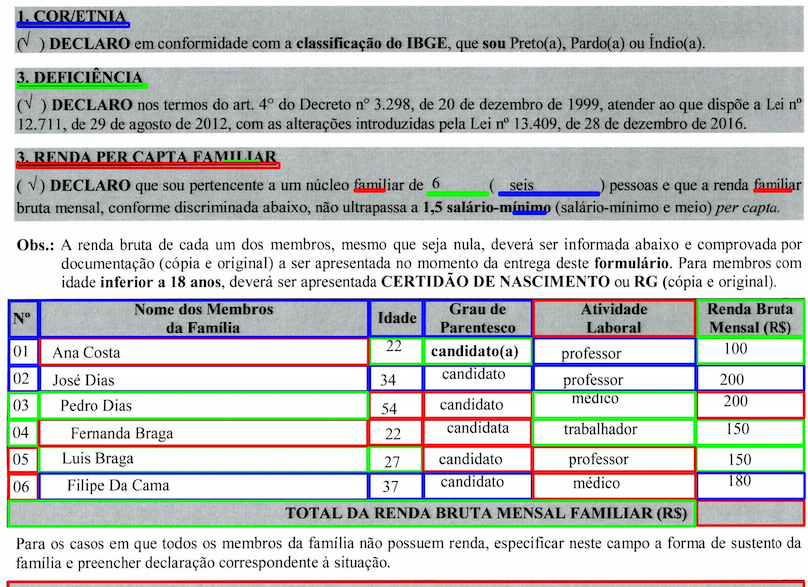}
    \caption{}
  \end{subfigure}
    \begin{subfigure}{0.45\linewidth}
  \centering
    \includegraphics[width=\linewidth]{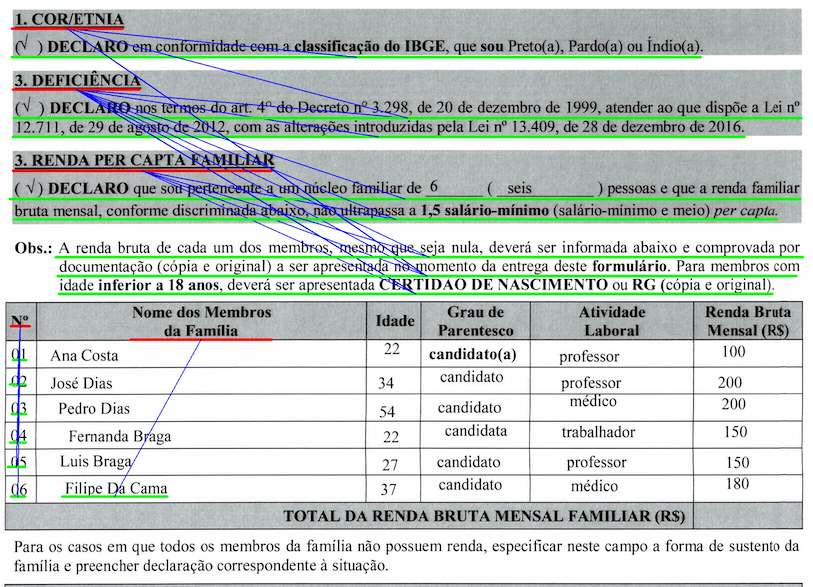}
    \caption{}
  \end{subfigure}
  \caption{Visualization of paragraph-level regions (a), tabular regions (b) and predictions (c) for a Portuguese form in XFUND.
  }
  \label{fig:abl_pt}
\end{figure*}

Figure \ref{fig:abl_pt} visualizes paragraph-level regions, tabular regions, and predictions for a Portuguese form in FUNSD. It shows that paragraph-level regions are suitable for the top portion of the form, while tabular regions specifically pertain to the bottom table. In this particular form, the decision was made to adopt paragraph-level regions, resulting in the exclusion of the tabular layout despite its ability to convey more information. 
We acknowledge that there are instances where our proposed approach may struggle to accurately distinguish between paragraph-level and tabular regions, leading to a performance decrease.

\section{Visualizations of Ablation Results}
\label{sec:viz_ablation}
\begin{figure*}[!htp]
  \centering
  \begin{subfigure}{0.4\linewidth}
    \includegraphics[width=\linewidth]{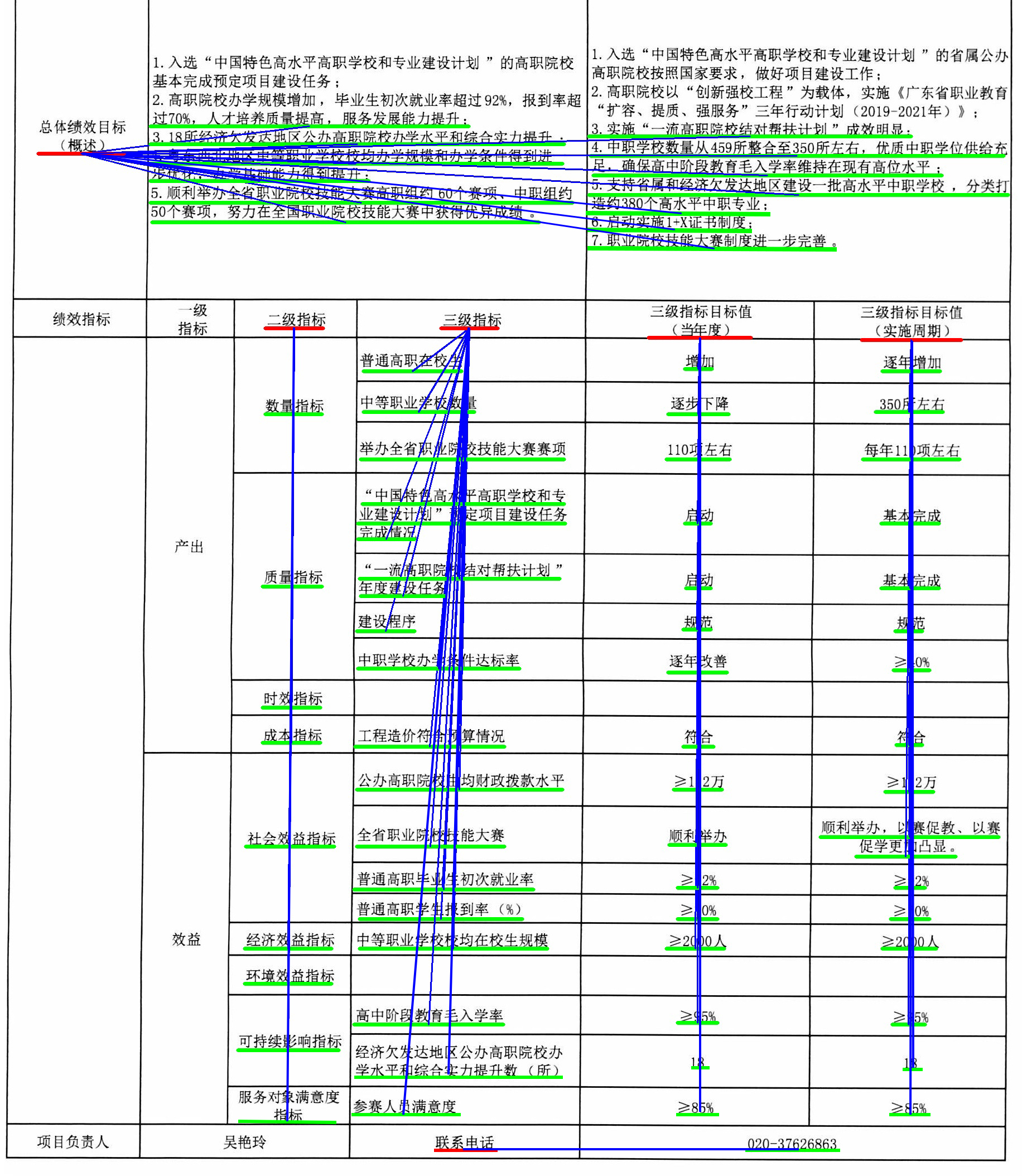}
    \caption{Ground Truth\\}
    \label{fig:subfig1}
  \end{subfigure}
  \begin{subfigure}{0.4\linewidth}
    \includegraphics[width=\linewidth]{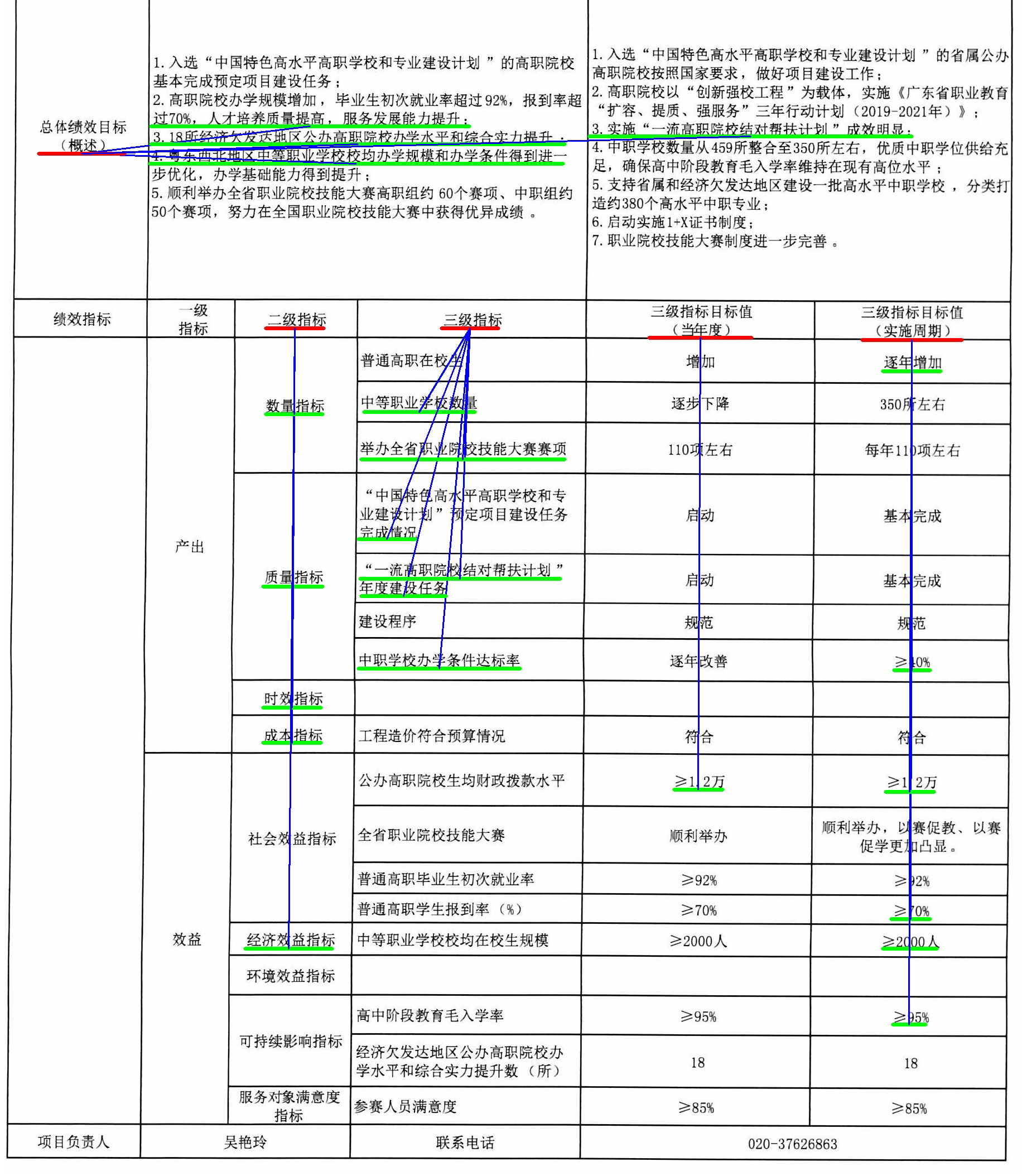}
    \caption{Concatenating only node embeddings\\}
    \label{fig:subfig2}
  \end{subfigure}
  \begin{subfigure}{0.4\linewidth}
    \includegraphics[width=\linewidth]{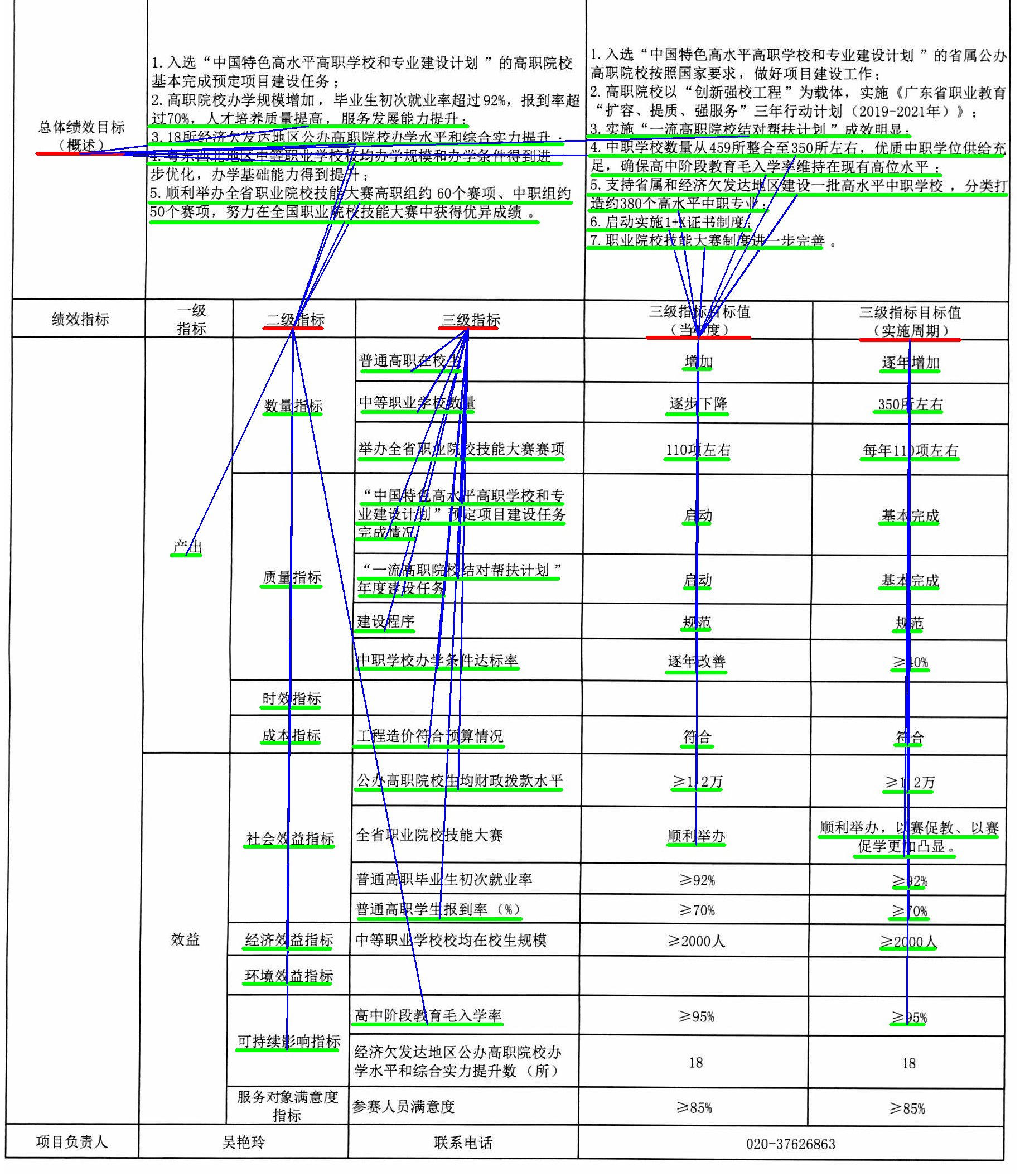}
    \caption{Concatenating only edge embeddings\\}
    \label{fig:subfig3}
  \end{subfigure}
  \begin{subfigure}{0.4\linewidth}
    \includegraphics[width=\linewidth]{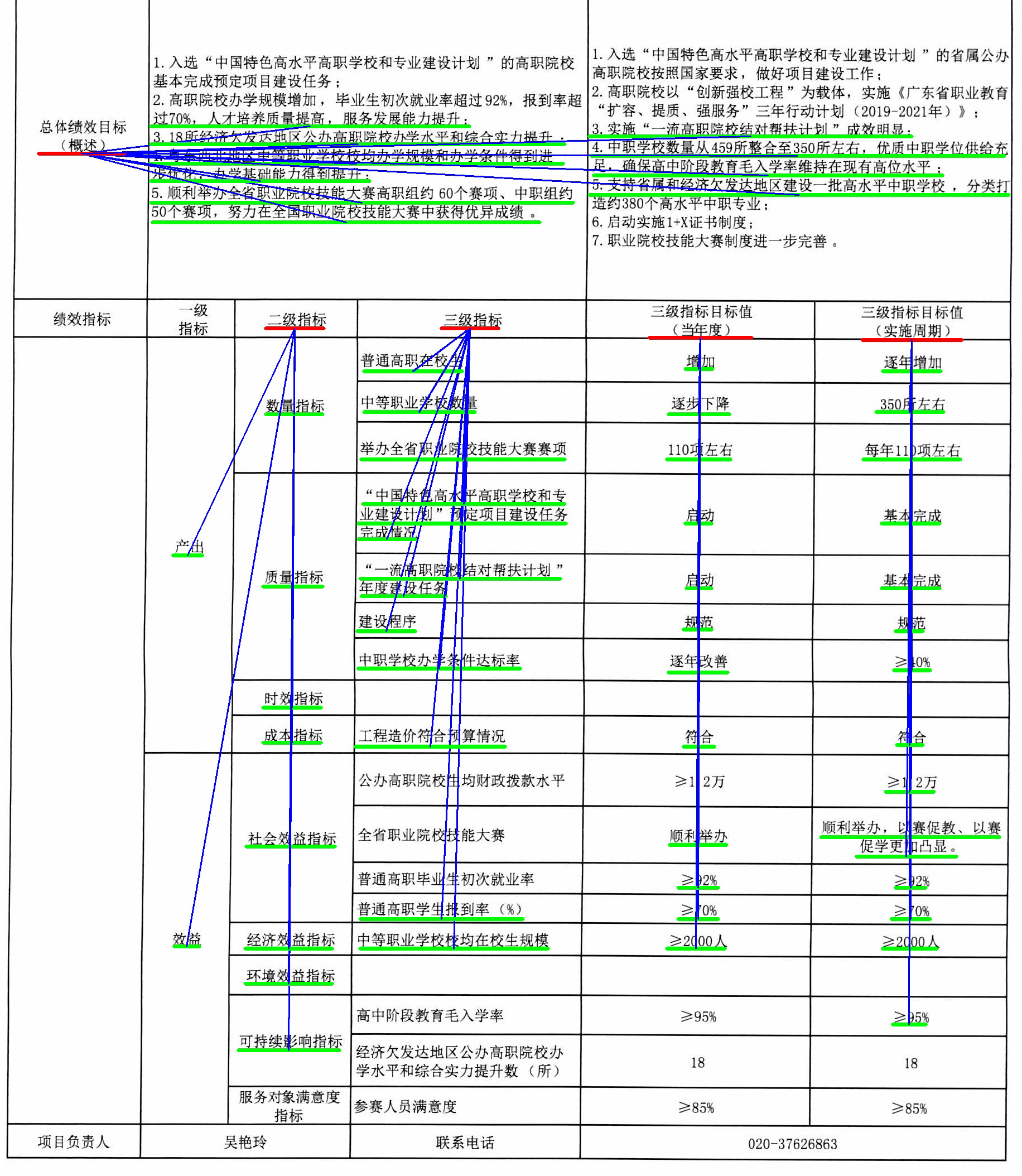}
    \caption{Concatenating node $\&$ edge embeddings
            }
    \label{fig:subfig4}
  \end{subfigure}
  \caption{Visualization of predictions of the ablation study of node and edge embeddings, 
  where \textcolor{red}{red} lines denote the question span, \textcolor{green}{green} lines denote the answer span, and \textcolor{blue}{blue} lines denote the question answer relation predictions.
  }
  \label{fig:abl1}
\end{figure*}

\begin{figure*}[!htp]
  \centering
  \begin{subfigure}{0.48\linewidth}
  \centering
    \includegraphics[width=0.8\linewidth]{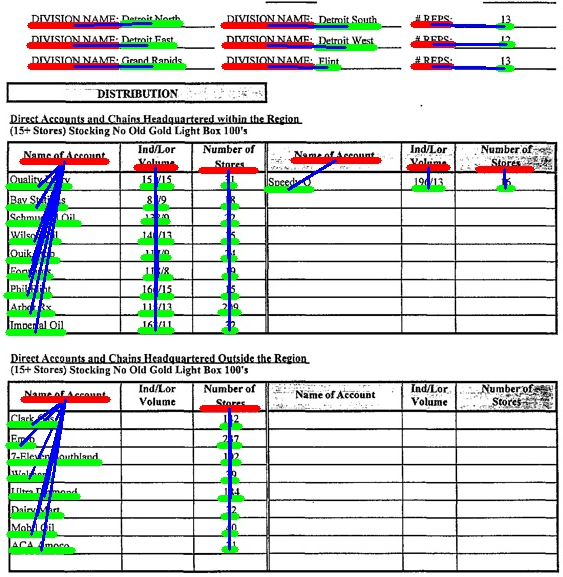}
    \caption{Ground Truth}
  \end{subfigure}
  \begin{subfigure}{0.48\linewidth}
  \centering
    \includegraphics[width=0.8\linewidth]{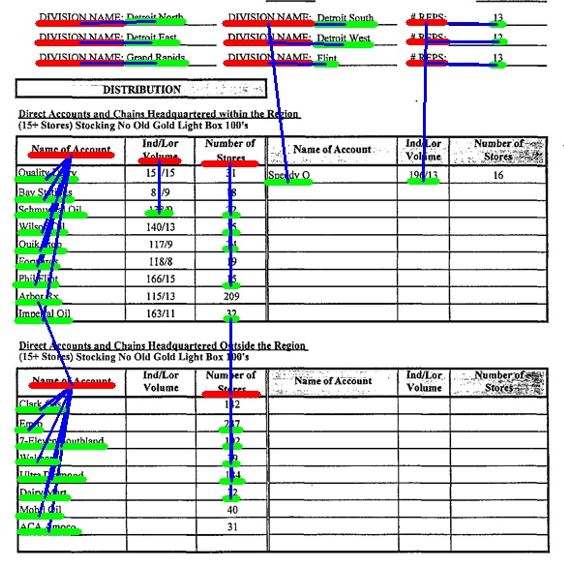}
    \caption{Predictions without paragraph/tabular region information}
  \end{subfigure}
  \begin{subfigure}{0.48\linewidth}
  \centering
    \includegraphics[width=0.8\linewidth]{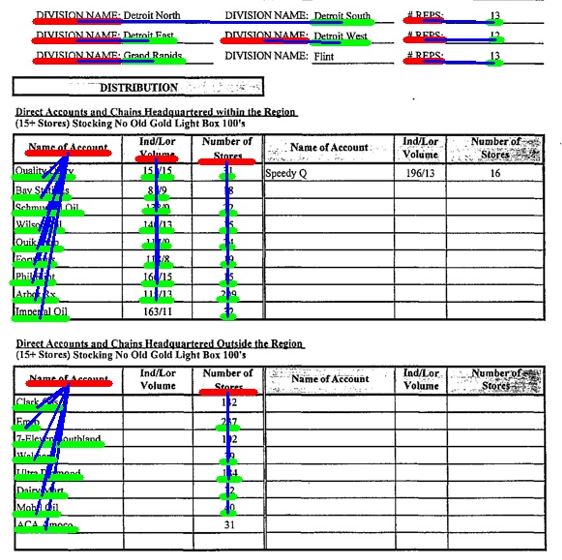}
    \caption{Predictions with paragraph/tabular region information}
  \end{subfigure}
  \caption{Visualization of predictions of the ablation study of incorporating paragraph/tabular regions.
  }
  \label{fig:abl2}
\end{figure*}

Figure \ref{fig:abl1} shows the visualization of predictions of the ablation study of node and edge embeddings. Figure \ref{fig:abl2} shows the visualization of predictions of the ablation study of incorporating paragraph/tabular regions.

\section{Pseudocode}
The following pseudocode extracts the tabular and paragraph-level regions from a VRD.
\label{sec:pseudocode}
\begin{algorithm*}
\caption{IdentifyHorizontalAndVerticalLines(image)}

    \begin{algorithmic}[1]
    \State Apply horizontal kernel to the image
    \State Apply vertical kernel to the image
    \State Find horizontal lines
    \State Find vertical lines   
    \State \Return Combined horizontal and vertical lines

\end{algorithmic}
\end{algorithm*}

\begin{algorithm*}
\caption{FindBoundingBoxes(lines)}
\begin{algorithmic}[1]

    \State TabularBoxList = []
    \State Find contours in lines
    \For{each contour}
        \State Compute the bounding box
        \State Append the box to TabularBoxList
    \EndFor
    \State \Return TabularBoxList
\end{algorithmic}
\end{algorithm*}

\begin{algorithm*}
\caption{SortBoxesByArea(boundingBoxes)}
\begin{algorithmic}[1]
    \State Sort the bounding boxes by area in increasing order
    \State \Return boundingBoxes
\end{algorithmic}
\end{algorithm*}

\begin{algorithm*}
\caption{AppendBoxToList(boundingBoxes, text)}
\begin{algorithmic}[1]
    \State FinalBoxList = []
    \For{each box in boundingBoxes}
        \If{the box contains any text and has no intersection with existing boxes in FinalBoxList}
            \State Append the box to FinalBoxList
        \EndIf
    \EndFor
    \State \Return FinalBoxList
\end{algorithmic}
\end{algorithm*}

\begin{algorithm*}
\caption{CheckAllTextPresent(FinalBoxList, text)}
\begin{algorithmic}[1]
    \If{all the text in the document is present in the boxes in FinalBoxList}
        \State \Return True
        \Else
    \State \Return False
    \EndIf
\end{algorithmic}
\end{algorithm*}

\begin{algorithm*}
\caption{GetMissingText(FinalBoxList, text)}

\begin{algorithmic}[1]
    \State missingText = []
    \If{the text is not present inside the bounding boxes of any of the FinalBoxList}
        \State Append text to missingText
    \EndIf
    \State \Return missingText
\end{algorithmic}
\end{algorithm*}

\begin{algorithm*}
\caption{AppendMissingTextBoxes(FinalBoxList, missingText, ParagraphRegions)}

\begin{algorithmic}[1]
    \For{each missing text in missingText}
        \If{missing text is present in any paragraph region in ParagraphRegions}
            \State Append paragraph region to FinalBoxList
        \EndIf
    \EndFor
    \State \Return FinalBoxList
\end{algorithmic}
\end{algorithm*}

\begin{algorithm*}
\begin{algorithmic}[1]
\caption{GetParagraphTabularRegions(imageFile)}
\State image = LoadImage(imageFile)
\State text = OCR(imageFile)
\State lines = IdentifyHorizontalAndVerticalLines(image)
\State boundingBoxes = FindBoundingBoxes(lines)
\State boundingBoxes = SortBoxesByArea(boundingBoxes)
\State FinalBoxList = AppendBoxToList(boundingBoxes, text)

\If{CheckAllTextPresent(FinalBoxList, text)}
    \State OutputResult(FinalBoxList)
\Else
    \State ParagraphRegions = GetParagraphRegionsFromEasyOCR(image)
    \State missingText = GetMissingText(FinalBoxList, text)
    \State FinalBoxList = AppendMissingTextBoxes(FinalBoxList, missingText, ParagraphRegions)
    \State OutputResult(FinalBoxList)
\EndIf
\end{algorithmic}
\end{algorithm*}

\section{Ablation Results of Constraint Loss}
The constraint loss has been modeled to encourage each answer entity to be linked to at most one question. Table \ref{tab:abl_constraint} shows that incorporating the constraint loss significantly improves the F1 score of \modelname, especially precision.
\label{sec:ablation}
\begin{table*}[!t]
\centering
\small
\begin{adjustbox}{width=\textwidth}
\begin{tabular}{c|c|c|c|c|c|c|c|c|c|c|c|c}
\toprule
\multirow{2}{*}{Model} & \multicolumn{3}{c|}{EN} & \multicolumn{3}{c|}{ZH} & \multicolumn{3}{c|}{JA} & \multicolumn{3}{c}{ES}\\ \cmidrule{2-13} 
& P & R & F1 & P & R & F1 & P & R & F1 & P & R & F1 \\ \midrule
\modelname & 69.71 & 73.74 & 71.67 & 76.80 & 82.62 & 79.60 & 70.16 & 81.39 & 75.36 & 70.26 & 81.78 & 75.59\\ \midrule
\modelname - constraint loss & 58.76 & 82.16 & 68.52 & 74.77 & 81.01 & 77.77 & 69.13 & 80.75 & 74.49 & 69.41 & 81.05 & 74.78\\ 
\end{tabular}
\end{adjustbox}
\begin{adjustbox}{width=\textwidth}
\begin{tabular}{c|c|c|c|c|c|c|c|c|c|c|c|c}
\midrule
\multirow{2}{*}{Model} & \multicolumn{3}{c|}{FR} & \multicolumn{3}{c|}{IT} & \multicolumn{3}{c|}{DE} & \multicolumn{3}{c}{PT}\\ \cmidrule{2-13} 
& P & R & F1 & P & R & F1 & P & R & F1 & P & R & F1 \\ \midrule
\modelname & 70.05 & 83.97 & 76.38 & 74.34 & 80.83 & 77.45 & 73.01 & 78.94 & 75.86 & 48.06 & 78.99 & 59.76\\ \midrule
\modelname - constraint loss & 71.54 & 80.57 & 75.79 & 72.32 & 79.32 & 75.66 & 71.74 & 75.59 & 73.61 & 46.98 & 74.02 & 57.48\\ \bottomrule
\end{tabular}
\end{adjustbox}
\caption{Precision, Recall and F1 score of ablation study of Constraint Loss on \modelname}
\label{tab:abl_constraint}
\end{table*}